\documentclass[10pt,twocolumn,letterpaper]{article}

\usepackage{cvpr}              

\usepackage{graphicx}
\usepackage{amsmath}
\usepackage{amssymb}
\usepackage{booktabs}
\usepackage{tabularx} 

\usepackage[pagebackref,breaklinks,colorlinks]{hyperref}

\usepackage[capitalize]{cleveref}
\crefname{section}{Sec.}{Secs.}
\Crefname{section}{Section}{Sections}
\Crefname{table}{Table}{Tables}
\crefname{table}{Tab.}{Tabs.}

\begin{document}

\title{A Comparative Study of Transfer Learning for Emotion Recognition using CNN and Modified VGG16 Models}

\author{Samay Nathani\\
Columbia University in the City of New York\\\\
{\tt\small sn3062@columbia.edu}
}
\maketitle

\begin{abstract}
Emotion recognition is a critical aspect of human interaction. This topic garnered significant attention in the field of artificial intelligence. In this study, we investigate the performance of convolutional neural network (CNN) and Modified VGG16 models for emotion recognition tasks across two datasets: FER2013 and AffectNet. Our aim is to measure the effectiveness of these models in identifying emotions and their ability to generalize to different and broader datasets. Our findings reveal that both models achieve reasonable performance on the FER2013 dataset, with the Modified VGG16 model demonstrating slightly increased accuracy. When evaluated on the AffectNet dataset, performance declines for both models, with the Modified VGG16 model continuing to outperform the CNN. Our study emphasizes the importance of dataset diversity in emotion recognition and discusses open problems and future research directions, including the exploration of multi-modal approaches and the development of more comprehensive datasets.
\end{abstract}

\section{Introduction}
\label{sec:intro}

Human emotion recognition based on facial expression plays an integral role in various applications within computing, such as human-computer interaction\cite{911197} and affective computing\cite{marin2018affective}, and outside of computing like mental health assessments\cite{turcian2023realtime}. However, accurate emotion recognition remains a challenging computer vision problem due to the complexity and variability of facial expressions, and also emotion classes are at the cusp of subjective and objective. Transfer learning is a strong candidate to address this challenge by using pre-trained deep learning models on larger and more diverse datasets to improve performance on target tasks with limited data. While previous research has explored transfer learning for emotion recognition using deep learning architectures like VGG16\cite{simonyan2014very}, there remain gaps in generalization for emotion recognition outside of a controlled setting. In this study, we aim to fill this gap by conducting a comprehensive analysis of transfer learning using a CNN based on the original VGG16 model and a modified VGG16 models for emotion recognition. We will investigate how these models perform when transferred from a source dataset (FER2013) to a target dataset (AffectNet), analyze their strengths and limitations, and identify strategies to improve their performance. By discussing the strengths and weaknesses of different transfer learning approaches, our project aims to provide useful information for improving emotion recognition and understanding the underlying structures of these models.

\section{Methodology}
\label{sec:formatting}

In this study, our hypothesis is that transfer learning using deep learning architectures can improve the accuracy of emotion recognition on target datasets with limited training data. To investigate this hypothesis, we utilize two widely used datasets: the Facial Expression Recognition 2013 (FER2013) dataset\cite{goodfellow2013challenges} as the source dataset and AffectNet\cite{mollahosseini2017affectnet} as the target dataset. The FER2013 dataset consists of grayscale images of facial expressions categorized into seven emotion classes: angry, disgust, fear, happy, sad, surprise, and neutral. On the other hand, modified version of AffectNet dataset provides more diverse examples for facial expressions, along with mixed backgrounds, angles, and people, making it a suitable target dataset for evaluating transfer learning approaches. We preprocess the image data by resizing them to 48x48 pixels and normalizing pixel values to the range [0, 1]. For our experiments, we select two deep learning architectures: Convolutional Neural Network (CNN) and a modified VGG16. 

The modified VGG16 model presented here is a modified version of the original VGG16 architecture, with several key differences aimed at enhancing its capacity and performance. 

The original VGG16 architecture comprises 13 convolutional and 3 fully connected layers. In contrast, the modified VGG16 model incorporates an additional convolutional layer in each convolutional block, resulting in a total of 16 convolutional layers. This increased depth allows the model to learn more intricate hierarchical representations of the input data, potentially capturing finer details and patterns.

The modified VGG16 model introduces an extra fully connected layer with 2048 neurons between the existing fully connected layers. This additional layer provides the model with more capacity to capture complex relationships in the learned features, potentially improving its discriminative power.

The fully connected layers in the modified VGG16 model contains a larger number of neurons compared to the original VGG16 architecture. Specifically, the first two fully connected layers have 4096 neurons each, while the additional layer has 2048 neurons. This increased number of neurons in the fully connected layers enables the model to learn more diverse and expressive representations from the convolutional feature maps.

A dropout rate of 0.25 is employed after each fully connected layer in the modified VGG16 model. Dropout regularization helps prevent overfitting by randomly dropping a fraction of the neurons during training, encouraging the network to learn more generalizable features. In the original VGG16 model, dropout values are 0.5 after each fully connected layer. Rather than remove it, its inclusion in the modified architecture reinforces the model's capacity to generalize effectively.

The modified VGG16 model incorporates a learning rate scheduler using the ReduceLROnPlateau callback. This dynamic adjustment of the learning rate during training helps fine-tune the optimization process, potentially leading to improved convergence and performance.

In summary, the modified VGG16 model distinguishes itself from the original VGG16 architecture through its increased depth, additional fully connected layer, larger number of neurons, dropout regularization, and learning rate scheduler. These modifications are designed to augment the model's capacity to learn intricate representations of the input data and enhance its performance in emotion recognition tasks. While the efficacy of these enhancements may vary depending on the specific dataset and task, we believe that they collectively contribute to the model's superiority over the original VGG16 architecture in capturing and discriminating emotional cues.

These models are pretrained on the FER2013 dataset and fine-tuned on the modified AffectNet dataset using transfer learning techniques. We evaluate the performance of each model based on standard metrics such as accuracy, precision, recall, and F1-score, which are commonly used in classification tasks to assess the model's ability to correctly classify emotions. We also incorporate predictive entropy as a performance metric. Predictive entropy measures uncertainty of the model’s predictions, and is calculated using  probability distribution of the predicted class probabilities. Entropy is directly correlated with uncertainty, and this metric is useful in gauging the model’s ability to predict and perform in ambiguous situations. Ultimately, this metric highlights not just the performance of the model, but rather its reliability and robustness. 

Access the notebook with the models and training data \href{https://colab.research.google.com/drive/1gkqYdxFKp2vuzf-biH8NoesnJVRcBhN_?usp=sharing}{here}.

\section{Experimental Results \& Discussion}

The performance metrics of the CNN and Modified VGG16 models are compared across two different datasets: FER2013 and AffectNet.

On the FER2013 Dataset, the CNN achieved an accuracy of 66.20\%, with precision, recall, and F1-score all hovering around the same range of approximately 66\%. The predictive entropy was measured at 0.3977. The modified VGG16 outperformed the CNN model slightly, with an accuracy of 67.43\% and similar precision, recall, and F1-score metrics. The predictive entropy increased to 0.5588, indicating slightly higher uncertainty in the model's predictions compared to the CNN.
The slight improvement in performance metrics, particularly accuracy, suggests that the modified VGG16 model may have learned more complex features and patterns from the FER2013 dataset compared to the CNN model. However, it is also possible that the increase in accuracy may be the result of overfitting.

On the AffectNet Dataset, the CNN experienced a decrease in accuracy to 41.43\%. Precision, recall and F1-score decreased significantly.
The Modified VGG16 showed similar trends to the CNN model, with an accuracy of 42.86\% and a decrease in recall and F1-score compared to the FER2013 evaluation.
The decline in performance metrics when transitioning from the FER2013 to the AffectNet dataset suggests that both models may struggle to generalize well to different datasets with varying characteristics and distributions of emotional expressions. However, the accuracy, precision, and recall scores were all greater with the modified VGG16 model, suggesting the more complex architecture played a role its higher performance on the target dataset. 

While the Modified VGG16 model demonstrated slightly superior performance compared to the CNN model on the FER2013 dataset, both models experienced a notable decrease in performance when evaluated on the AffectNet dataset. This highlights the importance of dataset diversity and the challenges of generalization in emotion recognition tasks. Further improvements in model architectures and training methodologies may be necessary to enhance performance across diverse datasets.

The CNN, known for its simplicity and computational efficiency, emerged as a baseline in our comparative analysis. Its architecture, characterized by alternating convolutional and pooling layers, demonstrated success in extracting spatial features from input images. As a result, the CNN exhibited modest yet robust performance in capturing the subtle nuances of emotional cues, achieving competitive accuracy and precision scores across multiple emotion classes well above random. 

In contrast, the modified VGG16 model has a slightly deeper and more complex architecture. The modified VGG16 model showcased slightly superior performance compared to the CNN, achieving greater accuracy, precision, and recall scores across a range of emotion classes.

However, the heightened performance of the modified VGG16 model came at the expense of increased computational complexity and resource requirements. Its more complex architecture, comprising multiple convolutional and fully connected layers, required longer training times, higher memory consumption, and overall more compute resources. As a result, the practical usage of the modified VGG16 model may be hindered by its computational demands, particularly in resource-constrained environments, all for marginally better performance when compared to the CNN.

Furthermore, our analysis revealed insights into the tradeoffs between model complexity and generalization. While the modified VGG16 model demonstrated notable proficiency with the training data, concerns regarding overfitting were observed, particularly when trained on smaller datasets. This highlights the importance of  regularization techniques and data augmentation strategies to mitigate the risk of overfitting.

\section{Conclusion}

This study investigated the performance of CNN and Modified VGG16 models for emotion recognition tasks across two datasets, FER2013 and AffectNet. We aimed to assess the effectiveness of these models in recognizing emotion and their ability to generalize to more diverse datasets.

Our findings revealed that both models achieved reasonable performance on the FER2013 dataset, with the Modified VGG16 model exhibiting slightly superior accuracy compared to the CNN model, which was based on the original VGG16 model. However, when evaluated on the AffectNet dataset, both models experienced a decrease in performance, which speaks to the challenges of generalization to new data.

A highlight from this study is the importance of dataset diversity in training and evaluating. The performance disparities observed between the FER2013 and AffectNet datasets emphasize the need for more comprehensive and representative datasets to improve the robustness and generalization capabilities of these models.

Despite the preliminary results achieved on the FER2013 dataset, this study calls out open problems and areas for future research. One such area is the exploration of multi-modal approaches to emotion recognition, which integrate information from facial images, text, and audio to enhance model performance.

Additionally, addressing the limitations of existing datasets and developing improved methods for data collection and annotation could further improve  accuracy and reliability. Furthermore, investigating the influence cultural differences and situational contexts could provide valuable insights into the complexities of human emotion recognition.

To conclude, our study contributes to the ongoing research in emotion recognition by highlighting strengths and limitations of CNN and Modified VGG16 models and by calling out the importance of dataset diversity and future research directions. By addressing these challenges and exploring new avenues for research, we can advance the field of emotion recognition and its applications in affective computing and human-computer interaction.

\begin{figure}[h] 
    \centering 
    \includegraphics[width=0.45\textwidth]{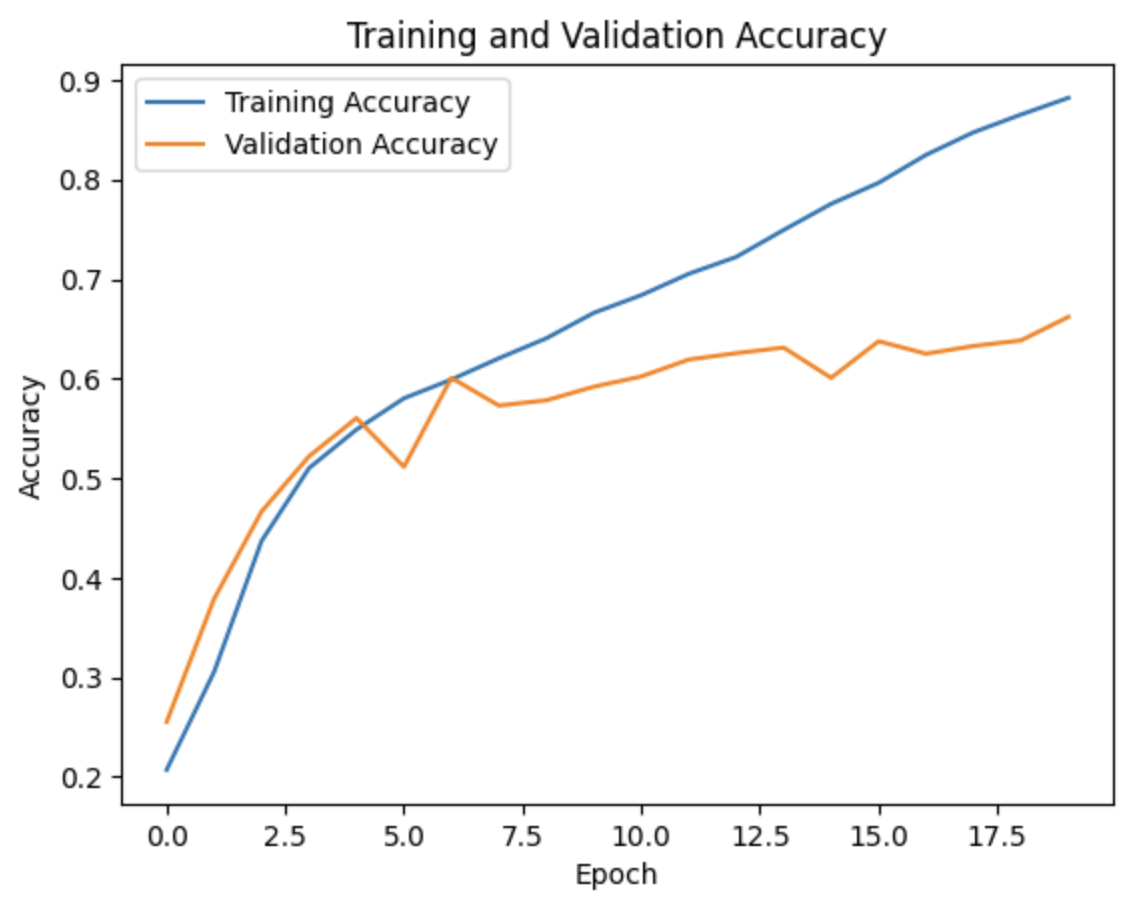} 
    \caption{CNN Accuracy on the FER2013 dataset} 
    \label{fig:example} 
\end{figure} 

\clearpage

\begin{table}[h]
\centering
\begin{tabular}{@{}lccccc@{}}
\toprule
\textbf{Model} & \textbf{Accuracy (\%)} & \textbf{Precision (\%)}\\ 
\midrule

CNN & 66.20 & 66.27 \\ 
Modified VGG16& 67.43 & 67.60 \\ \bottomrule

\end{tabular}

\caption{Accuracy and precision of CNN and Modified VGG16 models with the FER2013 Dataset.}
\label{tab:performance_metrics}
\end{table}

\begin{table}[h]
\centering
\begin{tabular}{@{}lccccc@{}}
\toprule
\textbf{Model} & \textbf{Recall (\%)} & \textbf{F1-Score (\%)} \\
\midrule
CNN & 66.20 & 66.08 \\ 
Modified VGG16 & 67.43 & 67.35 \\ \bottomrule
\end{tabular}

\caption{Recall and F-1 score of CNN and Modified VGG16 models with the FER2013 Dataset.}
\label{tab:performance_metrics}
\end{table}

\begin{table}[h]
\centering
\begin{tabular}{@{}lccccc@{}}
\toprule
\textbf{Model} & \textbf{Predictive Entropy}\\ \midrule

CNN & 0.3977 \\ 
Modified VGG16 & 0.5588 \\ \bottomrule

\end{tabular}

\caption{Predictive Entropy of CNN and Modified VGG16 models with the FER2013 Dataset.}
\label{tab:performance_metrics}
\end{table}

\begin{table}[h]
\centering
\begin{tabular}{@{}lcccc@{}}
\toprule
\textbf{Model} & \textbf{Accuracy (\%)} & \textbf{Precision (\%)} \\
\midrule
CNN & 41.43 & 51.34  \\
Modified VGG16 & 42.86 & 56.70 \\
\bottomrule
\end{tabular}
\caption{Accuracy and Precision of the CNN and Modified VGG16 models on the AffectNet Dataset.}
\label{tab:model_metrics}
\end{table}

\begin{table}[h]
\centering
\begin{tabular}{@{}lcccc@{}}
\toprule
\textbf{Model} & \textbf{Recall (\%)} & \textbf{F1-Score (\%)} \\
\midrule
CNN &  41.43 & 38.97 \\
Modified VGG16  & 42.86 & 38.81 \\
\bottomrule
\end{tabular}
\caption{Recall and F-1 Score of the CNN and Modified VGG16 models on the AffectNet Dataset.}
\label{tab:model_metrics}
\end{table}

\begin{figure}[h] 
    \centering 
    \includegraphics[width=0.45\textwidth]{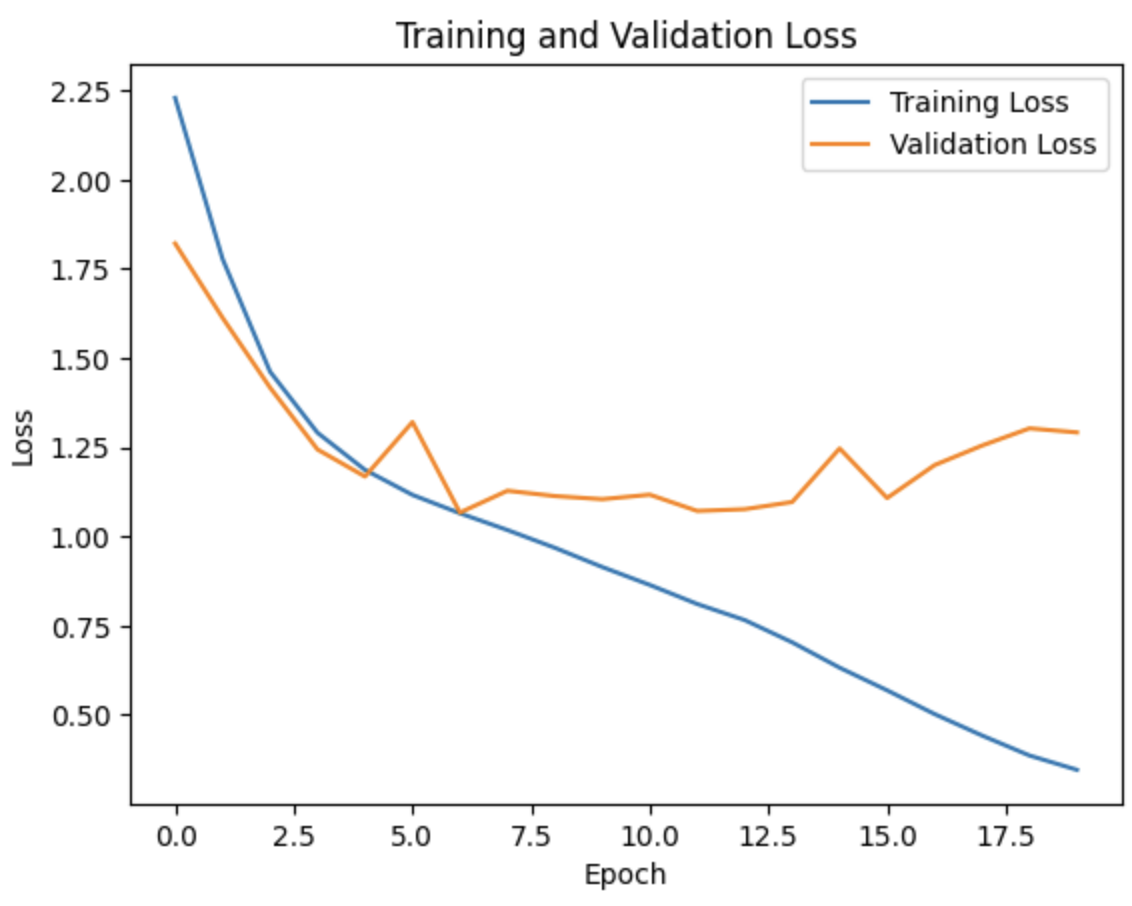} 
    \caption{CNN Loss on the FER2013 dataset} 
    \label{fig:example} 
\end{figure} 

\begin{figure}[h] 
    \centering 
    \includegraphics[width=0.45\textwidth]{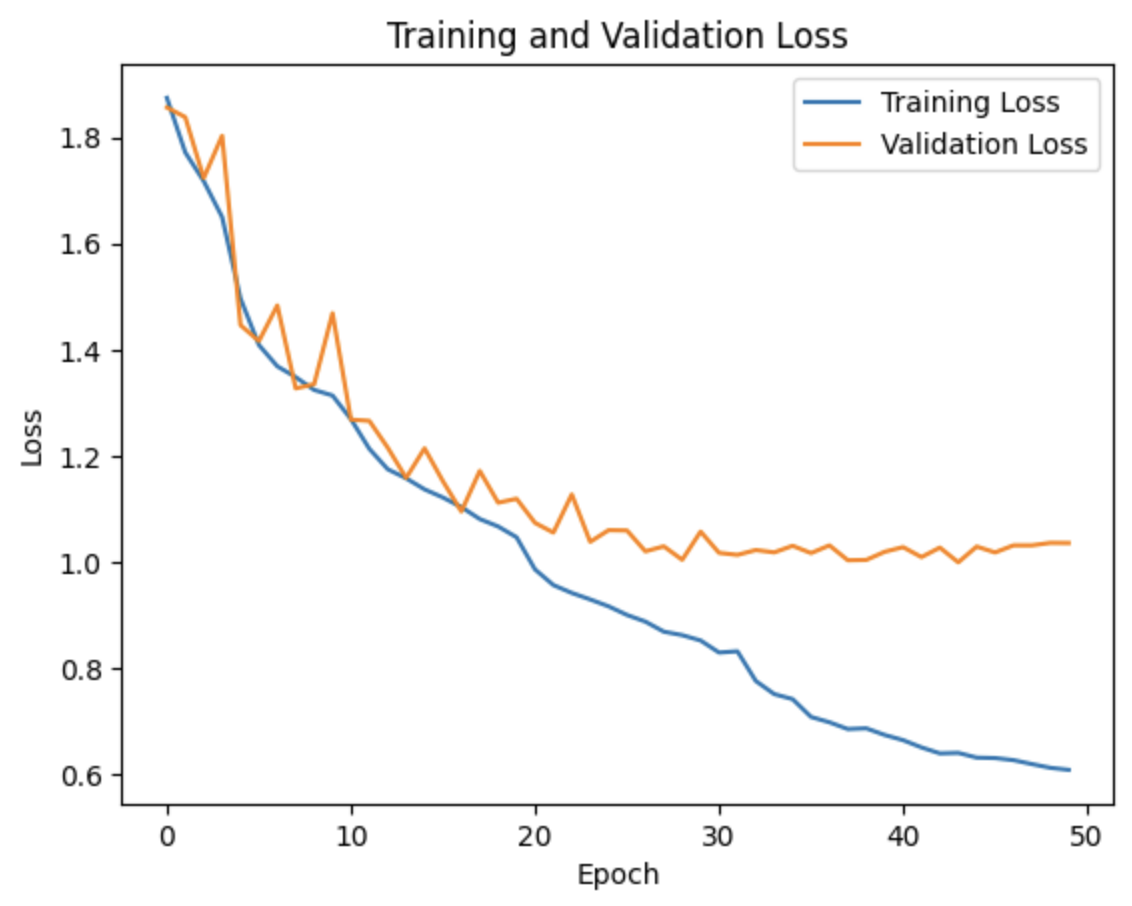} 
    \caption{Modified VGG16 Loss on the FER2013 dataset} 
    \label{fig:example} 
\end{figure} 

\begin{figure}[h] 
    \centering 
    \includegraphics[width=0.45\textwidth]{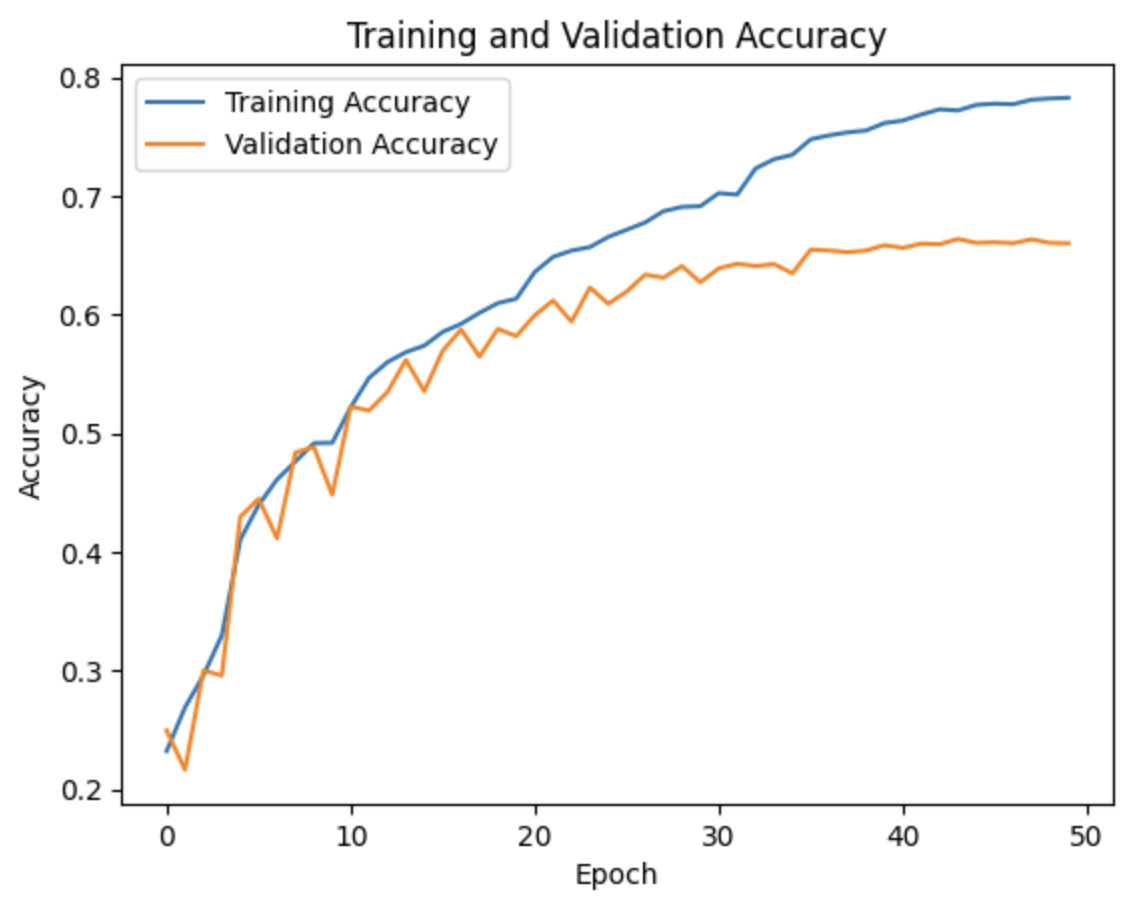} 
    \caption{Modified VGG16 Accuracy on the FER2013 dataset} 
    \label{fig:example} 
\end{figure} 

\clearpage

\bibliographystyle{ieee_fullname}
\bibliography{egbib}

\end{document}